\documentclass[10pt,twocolumn,letterpaper]{article}
\usepackage{times}
\usepackage{epsfig}
\usepackage{graphicx}
\usepackage{amsmath}
\usepackage{amssymb}
\usepackage{booktabs}
\usepackage{caption}
\usepackage{authblk}

\usepackage[breaklinks=true,bookmarks=false]{hyperref}

\begin{document}

\title{Blocks2World: Controlling Realistic Scenes with Editable Primitives}


\author[1]{Vaibhav Vavilala} 
\author[1,*]{Seemandhar Jain} 
\author[1,*]{Rahul Vasanth} 
\author[1]{Anand Bhattad} 
\author[1]{David Forsyth} 

\affil[1]{University of Illinois at Urbana-Champaign}
\affil[*]{Equal contribution}


\twocolumn[{%
\renewcommand\twocolumn[1][]{#1}%
\maketitle
\begin{center}
    \centering
    \captionsetup{type=figure}
    \includegraphics[width=\textwidth]{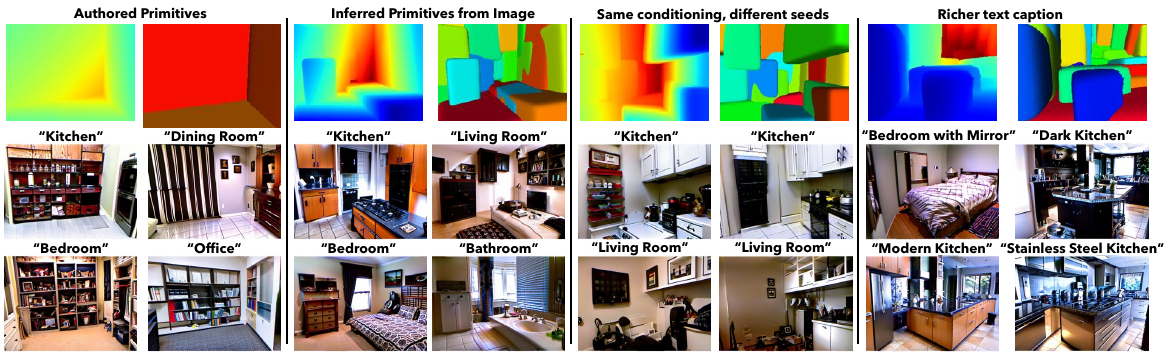}
     \captionof{figure}{\small{Blocks2World makes it possible to author indoor scenes using geometric primitive representations
    conditioning a diffusion model. The method can map authored
    primitives to images \textit{(first pair of columns)}. Using one set of primitives, we can create images with different scene labels
    {\em with the same geometry} \textit{(second pair)}. Alternatively, we can create multiple images of a given scene type
    \textit{(third pair)}.  By adding details to the text description of the desired scene, we can change its appearance
    \textit{(fourth pair)}. Other authoring activities, including moving primitives within a scene and moving the camera, are
    described below.}}
     \label{fig:teaser}
\end{center}%
}]



\begin{abstract} We present Blocks2World, a novel method for 3D scene rendering and editing that leverages a two-step process: convex decomposition of images and conditioned synthesis. Our technique begins by extracting 3D parallelepipeds from various objects in a given scene using convex decomposition, thus obtaining a primitive representation of the scene. These primitives are then utilized to generate paired data through simple ray-traced depth maps. The next stage involves training a conditioned model that learns to generate images from the 2D-rendered convex primitives. This step establishes a direct mapping between the 3D model and its 2D representation, effectively learning the transition from a 3D model to an image. Once the model is fully trained, it offers remarkable control over the synthesis of novel and edited scenes. This is achieved by manipulating the primitives at test time, including translating or adding them, thereby enabling a highly customizable scene rendering process. Our method provides a fresh perspective on 3D scene rendering and editing, offering control and flexibility. It opens up new avenues for research and applications in the field, including authoring and data augmentation.
\end{abstract}

\section{Introduction}
In a traditional graphics pipeline, objects are represented by textured geometry (primarily meshes) which are then rendered using physically-based methods, involving intricate compromises between speed and accuracy. While this approach provides precise control over the resulting images, creating objects can be difficult and costly. 

In this paper, we propose a fundamentally different and intuitive approach to scene rendering and editing - {\bf Blocks2World}. Blocks2World is built around a simple geometric interface to a diffusion model, enabling easier authoring while maintaining high-level geometric control. To date, diffusion models have
been controlled with retinotopic maps~\cite{zhang2023adding} or pixel movements ~\cite{endo2022user, pan2023drag}.  In
contrast,  Blocks2World allows an artist to either build up a
set of primitives or edit preexisting primitives automatically extracted from an image, and then combine them with text prompts. This combined input is passed to a statistical renderer — a diffusion model conditioned on primitives — which then generates a realistic scene.

Blocks2World provides an interactive and responsive interface, where artists can directly manipulate a primitive or camera, and the scene changes appropriately. Text modifications also yield corresponding changes in scenes. The statistical renderer takes care of surfacing and ensures the realism of the resulting image. The trade-off here is less detailed geometric control for much easier authoring.


Blocks2World is built on existing technologies in a novel way.  First, we use a variant of CVXNet~\cite{deng2020cvxnet,vavilala2023convex}
to represent scenes with an assembly of cuboids, yielding a collection of tuples (image, cuboids).  Next, we use
ControlNet~\cite{zhang2023adding} to condition a stable diffusion image generator~\cite{rombach2021highresolution} on
depth maps derived from the cuboids.  Finally, we take authored cuboids, compute a depth  map and feed that into the
conditional image generator.    Controlling stable diffusion directly with depth is known, but depth maps are hard to
edit whereas our assemblies of cuboids are easy to edit. 


Our contributions are 
\begin{itemize}
  \item We present Blocks2World, a novel method for high-level scene rendering and editing. Our approach combines convex
  decomposition and conditioned synthesis for intuitive scene manipulation.
  \item Blocks2World can produce images of different types of scenes which {\em share} a prescribed geometry.
\item Extensive qualitative and quantitative evaluation demonstrates that, with Blocks2World, the author ``gets what
  they asked for.''
\end{itemize}

\section{Related Work}
\label{related}
Our work intersects with several fast-moving research areas. We acknowledge the contributions of previous work in primitive image decomposition, image editing and scene rearrangement. 

\paragraph{Primitive Decomposition} Historically, the idea of decomposing scenes or objects into primitives for computer vision has been widely studied~\cite{roberts, binford71, biederman}. Such representations leverage parsimonious abstraction~\cite{Chen2019BSPNetGC} and facilitate natural segmentation~\cite{biederman,binford71,lego}. The challenge has been in selecting primitives that are easily inferred from image data~\cite{nevatia77,shgc} and that allow for simplified geometric reasoning~\cite{hebertponce}. 

While the early focus was on individual objects as opposed to scenes, more recent research incorporates powerful neural methods to predict the appropriate set of primitives from data~\cite{abstractionTulsiani17, Zou_2018_CVPR, liu2022towards}.  Despite their strengths, these approaches face challenges in producing varying numbers of primitives per scene~\cite{jang2017categorical}. Related to our work is ~\cite{abstractionTulsiani17} effectively using deep networks to parse shapes as unions of basic rectangular prisms, achieving consistent interpretations across different shapes. ~\cite{Zou_2018_CVPR} propose an algorithm that can predict the 3D room layout from a single image, generalizing to different types of images and room layouts. The decomposition of outdoor scenes is another significant area of focus~\cite{hoiem:2005:popup,hoiemijcv2007, s:gupta10}, as is the parsing of indoor scenes~\cite{hedauiccv2009, hedaueccv2010, hedaucvpr2012, stekovic2020general, liu2018planenet, Zou_2017_ICCV, Fouhey13, jiang2014finding, kluger2021cuboids}. The effectiveness of descent methods always depends heavily on the initial starting point~\cite{ransac, Kang2020ARO, Ramamonjisoa2022MonteBoxFinderDA,Hampali2021MonteCS}. 


Closely related to our work is CvxNet~\cite{deng2020cvxnet}, based on the idea of convex decomposition, which is a way of representing an object as a collection of convex polytopes. We leverage an advanced version of CvxNet (currently under review) to decompose scenes—rather than just objects—into a comprehensive set of primitives. The contribution of our prior work principally lies in the extraction of these primitives to generate paired primitives and image sequences, which form the groundwork for this paper. 

Building on these primitives, we train conditional image editing models, enabling us to synthesize complex scenes. Once trained, our Blocks2World method provides the flexibility to significantly customize scene rendering. This is achieved by manipulating primitives at test time, which includes their movement or transformation, facilitating the creation of novel scenes and refined edits. We further illustrate how Blocks2World can be used to author scenes on demand, reinforcing the method's versatility and potential for broad creative applications.

%

\begin{figure*}[t!]
  \includegraphics[width=0.95\linewidth]{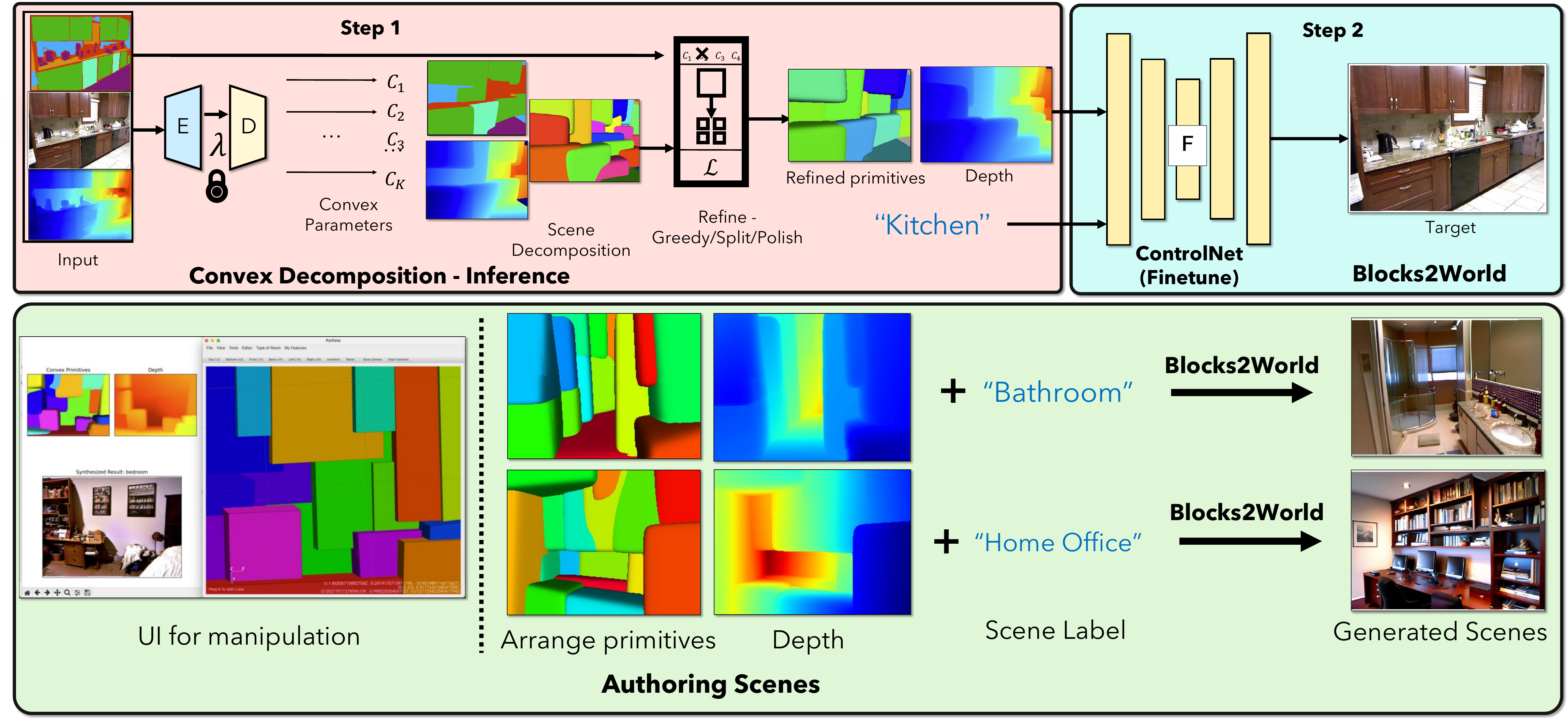}
  \caption{The Blocks2World Method: In the top row, we illustrate our two-step approach. Initially, a scene from NYUv2 is decomposed into 3D parallelepipeds (geometric primitives) using convex decomposition (Step 1). Following this, these primitives are ray-traced to generate pairs of natural images and primitive depth maps. A ControlNet is trained on these pairs (along with the scene label as the text prompt) to synthesize images from the primitive depth maps (Step 2).
  The result is a method that synthesizes images from depth maps, where
    the depth maps can be controlled by editing primitives.
    Finally, to author, we supply a set of primitives (authored, edited, or derived from an image; UI for manipulation in the bottom row, left); from this, we compute
    a depth map, pass this to our Blocks2World ControlNet with query terms, and generate an image (bottom row, right side).}
\label{fig:method}
\end{figure*}

\paragraph{Conditioned Image Synthesis} Seminal works such as Pix2Pix \cite{isola2017image} and CycleGAN \cite{zhu2017unpaired} have utilized Conditional Generative Adversarial Networks~\cite{goodfellow2014generative, mirza2014conditional} to demonstrate impressive translations between image domains, conditioned on source images. The work by \cite{chen2017photographic} further extended the idea of conditional synthesis to create photographic images from semantic layouts. For 3D scenes, the DeepSDF \cite{park2019deepsdf} leverages signed distance functions to generate 3D shapes. Most notably, the introduction of ControlNet \cite{zhang2023adding}, T2I-adaptor \cite{mou2023t2i} and StableDiffusion ~\cite{rombach2021highresolution} has improved conditional image synthesis quality significantly using diffusion-based models~\cite{dhariwal2021diffusion}. It allows users to add extra conditions to the diffusion model, such as the desired composition, pose, depth, or even color palette~\cite{vavilala2023applying}. Therefore, in this work we use ControlNet~\cite{zhang2023adding} to train a primitive-conditioned image generation diffusion model.

\paragraph{3D-aware image editing:} Recently, there has been substantial focus on integrating 3D awareness into image editing, primarily centered around individual objects rather than full scenes~\cite{gu2021stylenerf, chan2022efficient, wang2022score, poole2022dreamfusion, liu2023zero, tang2023make}. Our approach deviates from these in its comprehensive scene analysis, which extends beyond isolated objects. We decompose the entire scene into simple convex primitives for subsequent manipulation. From there, we synthesize new images conditioned on these primitives. An additional point of divergence from prior methods is our unique ability to author new images via the addition or editing of primitives ``authoring''), akin to assembling Lego pieces.

\paragraph{Scene rearrangement:} Current methodologies for scene rearrangement mainly revolve around robotic manipulation and embodied agents~\cite{king2016rearrangement, qureshi2021nerp, radford2021learning, liu2022structformer, wang2020scenem, weihs2021visual, wei2023lego}, seeking more intuitive and human-like commands for scene manipulation and navigation. Our work introduces a new perspective, focusing on object-level rearrangements within a single 2D image.

\section{Methodology}
In this section, we present the procedure for extracting primitives from an RGB image using our convex decomposition method as well as training the ControlNet conditioned on the depth of our primitives. In our results (Sec.~\ref{results}), we show several qualitative and quantitative evaluations of depth and scene label (addressing the question - did the user get what they asked for?).

\subsection{Choosing Conditioning Variables}

ControlNet~\cite{zhang2023adding} has been shown to be able to synthesize images based on depth maps; normal maps;
various edge maps; and various segment maps (at least!~\footnote{https://github.com/lllyasviel/ControlNet} offers
a wide range of cases). None is an appealing method to author images, because editing (say) a depth map or an edge map is complex. Instead, we want to control synthesized images with easily manipulated primitives.
Reliably fitting primitives to scene images remains hard (Sec.~\ref{related} and below), but we have
access to a method that can accurately fit smoothed, aligned cuboids (below). This method has eccentricities -- it tends to ``chatter'' by generating several primitives when fitting to walls, for example (as in
Figure~\ref{fig:figA}) -- but produces good fits. Representing a set of primitives presents challenges, so we
communicate with ControlNet with the depth map derived from the primitives.  In turn, this means that ControlNet must be
finetuned to cope with our depth maps.  As Fig.~\ref{fig:author} shows, Blocks2World is not distracted by the {\em absence} of
chatter, and our method can synthesize images from authored primitives.

Natural extensions involve conditioning ControlNet on normals or edges of the primitives. Initial experiments discourage this approach,
because normal effects and edge effects caused by chatter tend to force ControlNet to produce overly busy textures or
distracting small image features. 

\subsection{Convex Decomposition for Primitive Extraction}
Our first step is to use our previous work (under review and details in supplement) on convex decomposition to extract primitives from an RGB image. This is an extremely difficult optimization problem that we showed can be solved quite well with a mixed procedure - regression to get a starting point followed by optimization to polish the prediction significantly. This process yields 3D parallelepipeds that serve as basic building blocks for our model. Our training data consists of RGBD images, and we generate labeled samples from the depth map (1 for "inside", 0 for "outside") by defining boundaries near the surface of the depth map and at the top, bottom, left, right and back sides of it (as if it were a volume). Our
network predicts a fixed number of convexes (24 performed best on the NYUv2 dataset) that attempt to classify all points correctly (all inside points should be inside a convex; all outside points should be outside all convexes). A series of losses encourage stable training and well-organized primitives. From there, we implement a polishing step that takes the network prediction and refines the fit of the convexes, while removing unnecessary ones. Our method captures the geometric layout of the input quite well, as measured by traditional depth and normal error metrics, and works well even when the depth map is inferred by a depth estimation network e.g. \cite{Ranftl2021,bhat2023zoedepth}. See the top-left box of Fig. \ref{fig:method} for an overview of our primitive decomposition method.

\begin{figure*}[t!]
  \includegraphics[width=\textwidth]{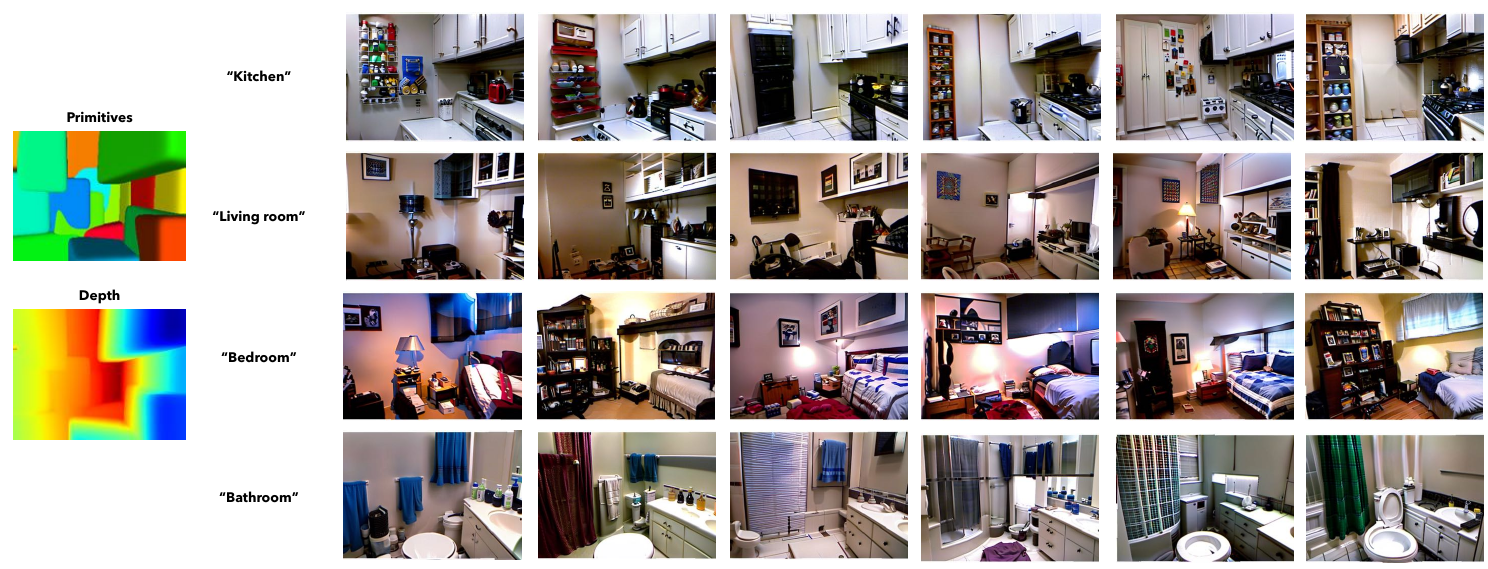}
  \caption{The diffusion model can generate many different {\tt kitchen}s (say) with a given geometry,
    using different random seeds.  This applies across many different scene descriptors, too.  Note how each synthesized
  image has geometry consistent with the query on the {\bf left}.}
\label{fig:C_seed}
\end{figure*}

\begin{figure*}[t!]
  \includegraphics[width=\textwidth]{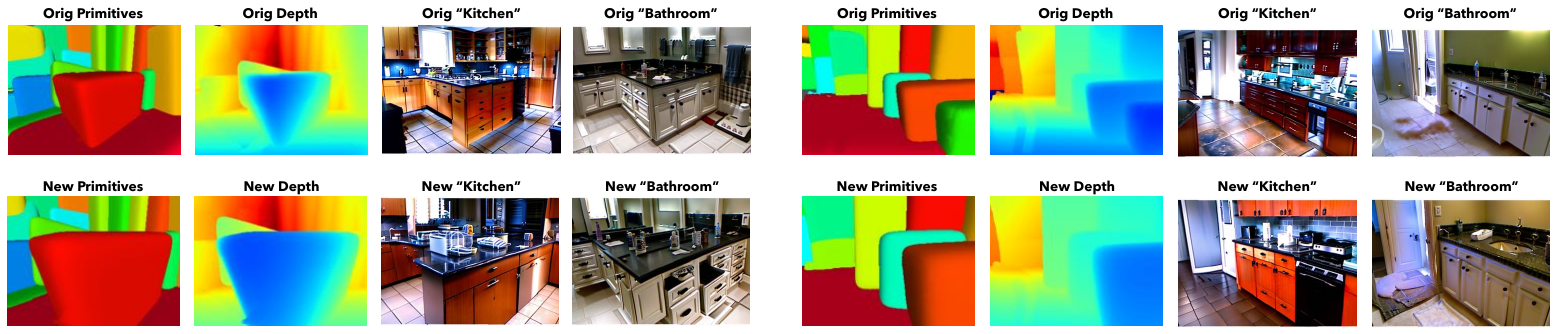}
  \caption{Our primitives are a genuine 3D representation of the desired scene. Moving the camera w.r.t. the primitives results in the appropriate change in the synthesized image.}
\label{fig:camMove}
\end{figure*}

\subsection{Training the Conditioned ControlNet}
Next, we train a ControlNet, based on the StableDiffusion 2.1 framework, conditioned on the depth of our primitives 
(overview: Fig. \ref{fig:method}). We use the 1449 NYUv2 images, generating primitives for each and raytracing them to obtain depth maps. Optionally, we also condition on a texture badge -- a hint consisting of some blocks of pixels of the target image removed corresponding to primitives that the ControlNet must inpaint (see Fig. \ref{fig:texture}). This process is
particularly useful for primitive editing, where we aim to remove, translate, or add a primitive to the scene while
preserving the lighting and texture of non-affected regions of the image. We train for 1200 epochs - 8 days on an Nvidia
A100 GPU, batch size 12, learning rate $3e-5$, locked diffusion model, mixed-precision training, and default
hyperparameters otherwise. The typical inference time for one image on one GPU at $512\times704$ resolution is under $6$ seconds.

\section{Results}
\label{results}
\subsection{Qualitative Evaluation}

\begin{figure*}
  \includegraphics[width=\textwidth]{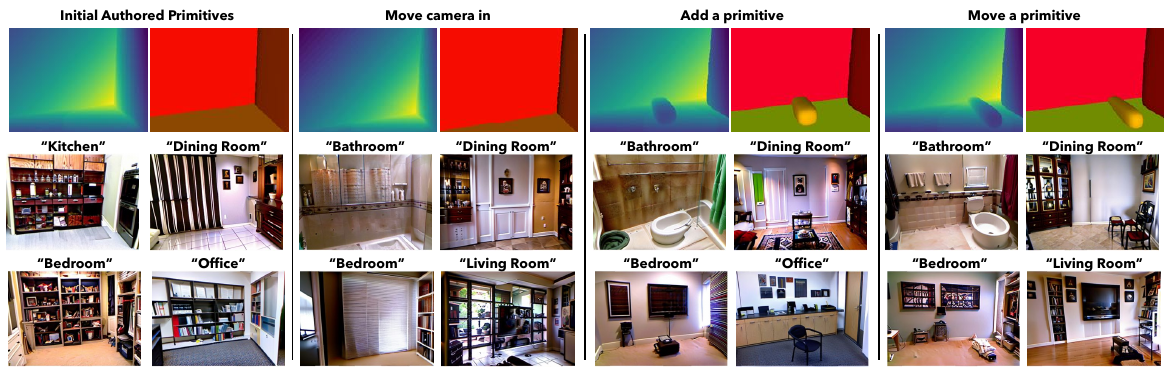}
  \caption{For most examples, we show queries using primitives derived from images, because they are easily available in quantity.  However, our method does not require primitives generated from images (which typically ``stutter'' on walls) for
    synthesis. The conditioning process is robust enough that we can synthesize images from authored primitives. 
    {\bf Top row:} shows depth (cool colors) and primitives (warm colors) for a set of authored
    primitives.  {\bf Bottom two rows:} Results for various query terms; note that editing the primitive layout has natural
    results.  
  }
\label{fig:author}
\end{figure*}

\begin{figure*}
  \includegraphics[width=1\textwidth]{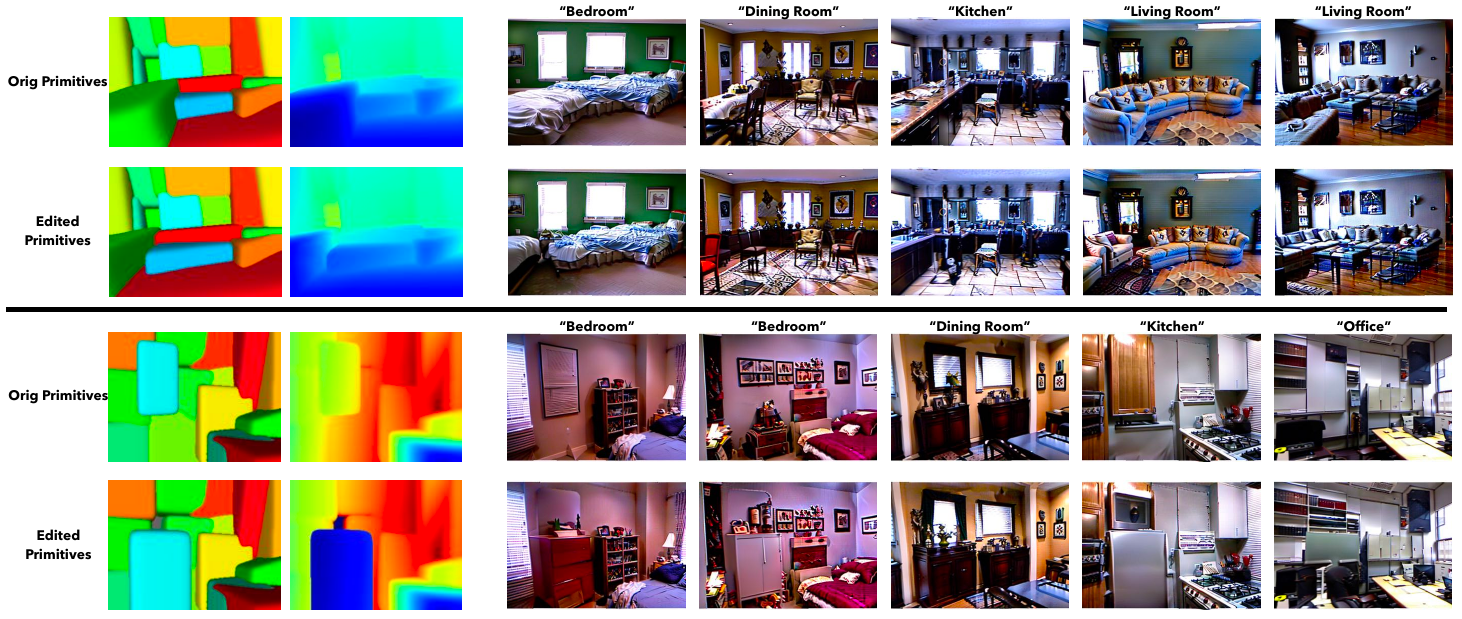}
  \caption{Moving a primitive causes the geometry of the synthesized scene to change appropriately.  We enable object moves without texture changes by masking off the primitive that is to be  moved in its source and target location and then constraining the method to fix
    pixels outside the mask.  In the {\bf top row}, the green cuboid on the bottom left is moved left, resulting in various furniture rearrangements to remain consistent with the geometry and maintain texture outside the masked region of the translated primitive. In the {\bf bottom row}, the blue cuboid on the top left is moved forward, with similar outcomes.}
\label{fig:movePrim}
\end{figure*}

\begin{figure*}
  \includegraphics[width=\textwidth]{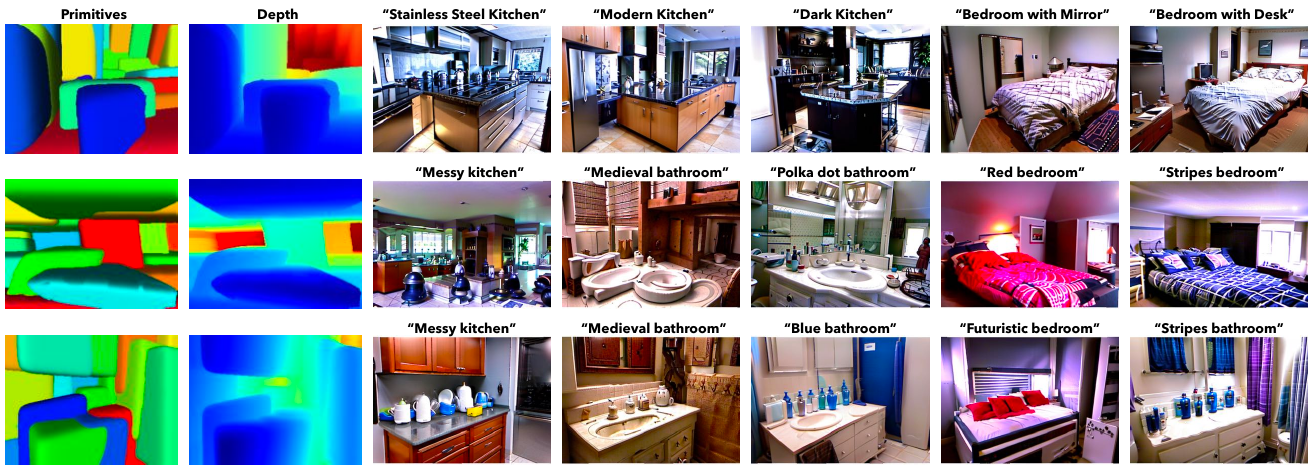}
  \caption{Although our rendering engine was only finetuned with simple scene labels (like "kitchen" or "bathroom"), we show here that text descriptions can be enriched with qualifiers, which the renderer respects (because the pretrained diffusion model was exposed to rich text caption - image pairs). Five synthesized images per row (with corresponding text caption) are shown alongside the primitives and depth conditioning (first two images per row).}
\label{fig:textLabel}
\end{figure*}

We perform extensive qualitative evaluations of our method. Firstly, we generate several scenes for the same geometry and scene label, varying the seed in each instance in Fig \ref{fig:C_seed}. Notice how the query geometry and scene label are respected while obtaining sample diversity with each different seed. Secondly, in Figs. \ref{fig:camMove}, \ref{fig:cam2}, we show that our synthesizer geometrically respects camera moves, demonstrating that our representation captures the high-level 3D layout of the scene. From there we examine our model from an artist's point of view: what if a user authors primitives? We show in Figs~\ref{fig:author},\ref{fig:author2} that this works as expected on scenes with small numbers of primitives i.e. our model doesn't require the dozen-plus primitives we generate automatically via our convex decomposition procedure. We built a simple UI to author primitives. Additional examples showing geometric and text-label consistency are shown in Fig.~\ref{fig:figA}. Further, as we show in Fig. \ref{fig:textLabel}, we can enrich the text label (e.g. \texttt{bedroom with desk}) - a key selling point of Stable Diffusion - with fairly high quality results. 


Next, a key question for artists is how well the synthesized image respects texture and lighting consistency after moving a primitive. We show our model can do this quite well in Fig.~\ref{fig:movePrim}. Regions corresponding to a shifted primitive are masked in the texture hint (see Fig.~\ref{fig:texture}) for the diffusion model to inpaint. The resulting image respects the geometry of the moved primitive and generally matches the lighting and texture of the source image.

\begin{figure*}
  \includegraphics[width=\textwidth]{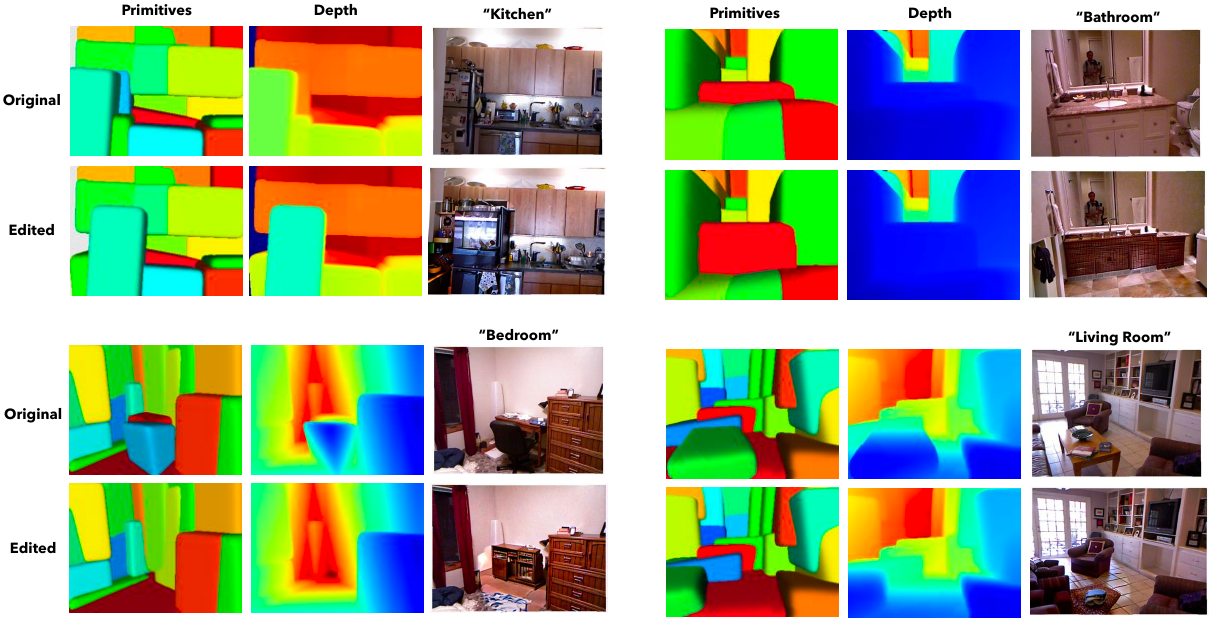}
  \caption{Blocks2World can edit real images. Given a real image, our convex decomposition method extracts primitives. Moving those primitives around results in geometrically sensible outputs while the unaffected regions of the image respect the original image lighting, texture, and geometry. In each set, the top row is a real scene with its primitives and depth. The bottom row shows edited primitives and the resulting edited image.}
\label{fig:editReal}
\end{figure*}

Finally, we can edit real images by first extracting the primitives, moving them around, and having Blocks2World synthesize a new image that geometrically aligns with the modified primitives, while maintaining the texture and lighting of the unaffected regions of the source image. We show this in Fig.~\ref{fig:editReal}.

\subsection{Quantitative Evaluation}
Our quantitative evaluation primarily focuses on verifying that the network generates the images we requested. This involves evaluating the scene classification and ensuring the depth of the output matches the input.

 \begin{figure}
  \includegraphics[width=0.9\linewidth]{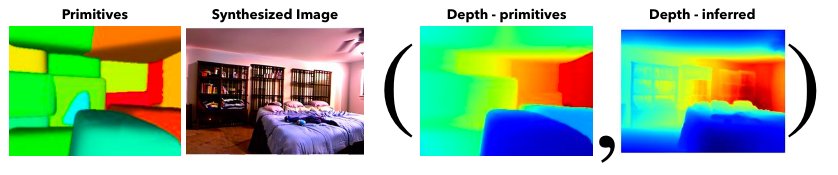}
  \caption{We can quantitatively evaluate depth by comparing the depth from primitives with the inferred depth (we use ZoeDepth-NK ~\cite{bhat2023zoedepth}) of the corresponding synthesized image.}
\label{fig:depth_eval}
\end{figure}

\begin{table}
\begin{center}
\resizebox{0.75\linewidth}{!}{%
\begin{tabular}{ c  c  c  c } 
 \toprule
 Cfg. & AbsRel & RMSE & RMSLE \\ [0.5ex] 
 \midrule
 bedroom & $0.131$ & $0.441$ & $0.115$ \\ 
 kitchen & $0.137$ & $0.476$ & $0.122$ \\
  
 living room & $0.139$ & $0.451$ & $0.121$ \\
 bathroom & $0.156$ & $0.537$ & $0.135$ \\
 dining room & $0.141$ & $0.473$ & $0.124$ \\
 office & $0.135$ & $0.460$ & $0.121$ \\
\midrule
 Avg. & $0.140$ & $0.473$ & $0.123$ \\
\midrule
 ZoeDepth \cite{bhat2023zoedepth} & \textbf{0.077} & \textbf{0.277} & \textbf{0.033} \\
 MIDAS \cite{Ranftl2021} & $0.110$ & $0.357$ & $0.045$ \\ 
\bottomrule

\end{tabular}
}
\caption{We evaluate whether our rendered image is consistent with demand by comparing depth computed from the synthesized image (inferred via~\cite{bhat2023zoedepth}) with demanded depth (primitives). Despite how simple the primitives are (and complex the scenes are), the overall depth error across the six most common classes is fairly low, comparable with recent single image depth estimation networks.}
 \label{tab:depth_err}
\end{center}
\end{table}
\textbf{Depth evaluation} For the primitives extracted from each of the 1449 NYUv2 images, we render one image each from the following set of six most common scene labels: \texttt{bedroom, kitchen, living room, bathroom, dining room, office}. As shown in Fig. \ref{fig:depth_eval}, we then compute depth error metrics for each synthesized image. We infer the depth from the synthetic image via ZoeDepth ~\cite{bhat2023zoedepth}, and use the primitive depth as the ground truth reference. We fit the unknown scale and shift parameters from the inferred depth map to the ground truth reference to improve the estimate. We show the error metrics in Table. \ref{tab:depth_err}, which demonstrate that the synthesized images respect the primitive depth very well, comparable in accuracy to recent SOTA depth prediction networks. 

\begin{figure}
\centering
 \includegraphics[scale=0.45]{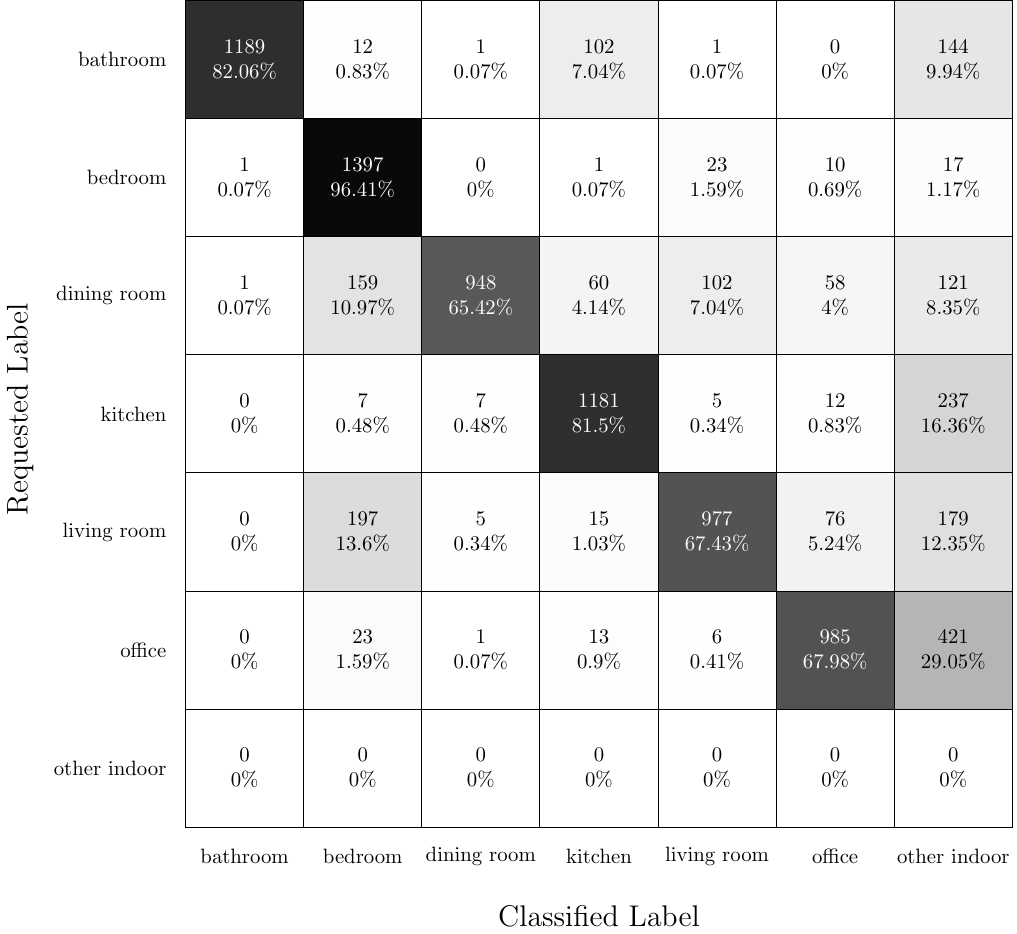}
\caption{We evaluate whether a user gets what they asked for by classifying the synthesized image with a SOTA scene classifier \protect \cite{seichter2022efficient}. The class-confusion matrix (showing the requested label against the label predicted for a synthesized image) suggests that our method produces the scene the user asks for. The error rate of our method (bAcc of $76.80$) is approximately that of the 
scene classifier (bAcc of $76.46$). Some scenes are more difficult than others, which is consistent with classifier behavior but may also result from the presence of objects that are poorly represented by our primitives, for example, chairs with legs in dining rooms.}
\label{table:ccm}
\end{figure}

\textbf{Scene label evaluation}
A key question for the user is - if I request a bedroom, did I get one? We can use a pretrained scene classification network ~\cite{seichter2022efficient} to compare the requested label with what the classification network predicts. As shown in Fig. ~\ref{table:ccm}, Blocks2World consistently generates images that match the requested scene label, achieving a bAcc of $76.80$. Note that this closely matches the classifier bAcc of $76.46$. We use the same dataset to evaluate as in our depth evaluation.


In our methodology, some errors can be accounted for due to a distribution shift between the source and generated datasets. We can quantify this difference via FID, obtaining 27.79 when comparing the 8,694 generated images and 1,145 ground truth NYUv2 images from the top 6 scene labels.





\section{Limitations and Conclusion}
We have introduced {\bf Blocks2World}, an intuitive approach to scene rendering and editing. Utilizing a two-step process of convex decomposition and conditioned synthesis, we convert simple block primitives into realistic images. This system offers user-friendly scene control and opens new opportunities in research and practical applications for scene authoring.

\begin{figure}[t!]
  \includegraphics[width=\linewidth]{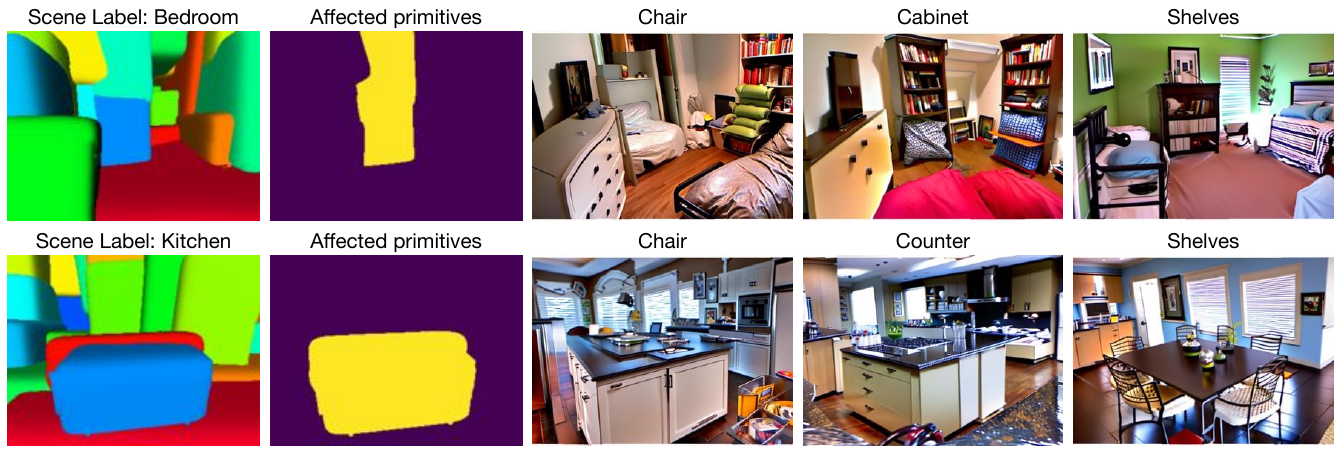}
  \caption{First row: synthesizing three images conditioned on the shown primitives and scene label``bedroom". We also condition the labeled primitives (in yellow) according to the class label shown in the rightmost three images. Second row: same thing conditioned on scene label ``kitchen." Notice improper registration between segmentation labels and primitives. Our model was unable to localize at times, or unable to even generate the object. We note that by trying several random seeds, a user may eventually get what was asked for, but our experimentation showed that proper object generation and localization were not consistent enough to quantitatively evaluate. We trained a Blocks2World model on segmentation labels by labeling each primitive with the most common ground truth segmentation label in its support.}
\label{fig:limit_a}
\end{figure}

\begin{figure*}
  \includegraphics[width=\textwidth]{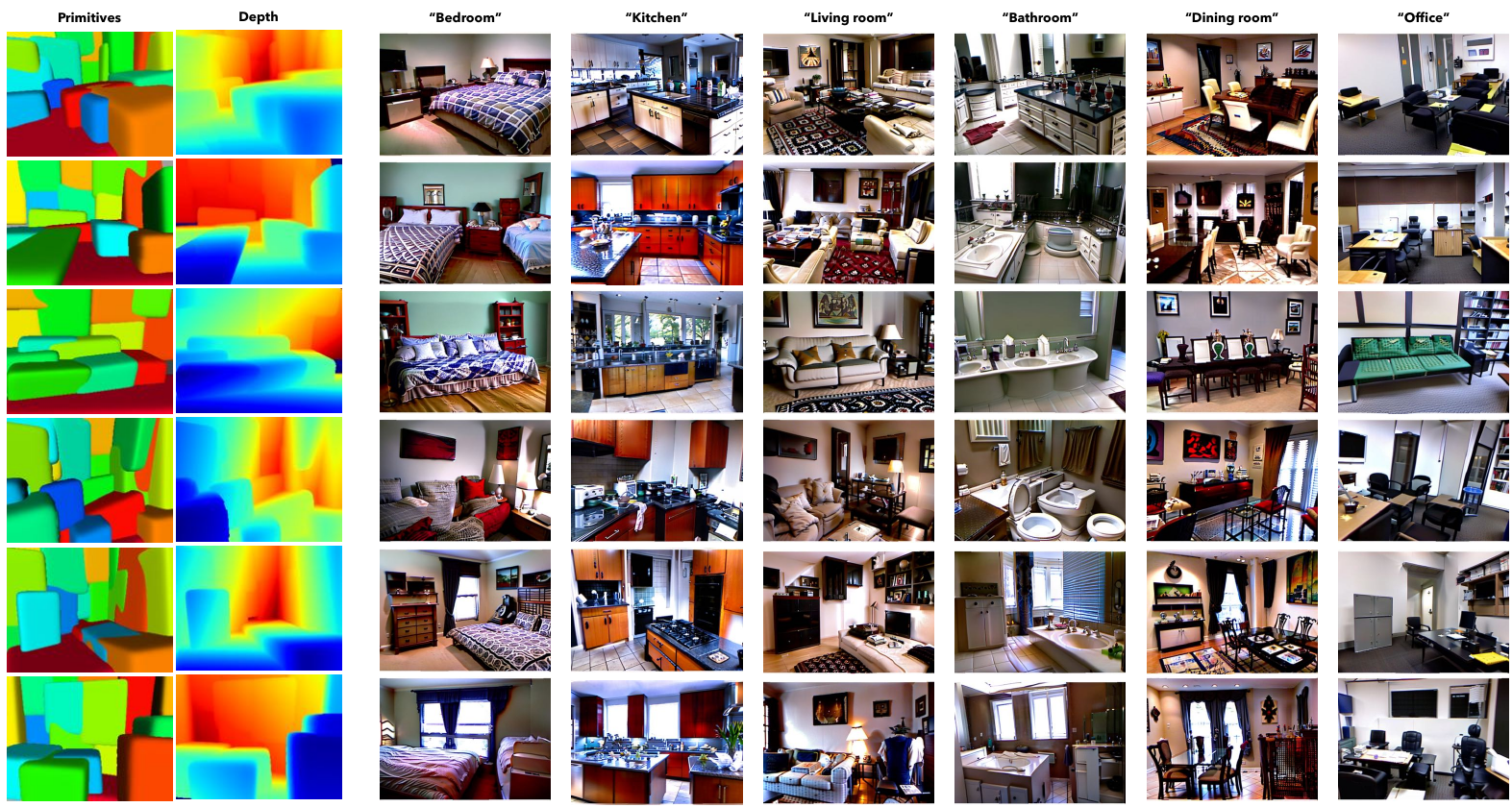}
  \vspace{-20pt}
  \caption{Further examples, showing query primitives (derived from images) and depth (from the primitives) on the {\bf
      left}, and various synthesized scene types on the {\bf right}.  Note how the synthesized image is controlled
    by the primitive geometry and by the scene label.}
\label{fig:figA}
\end{figure*}

\begin{figure}
  \includegraphics[width=0.95\linewidth]{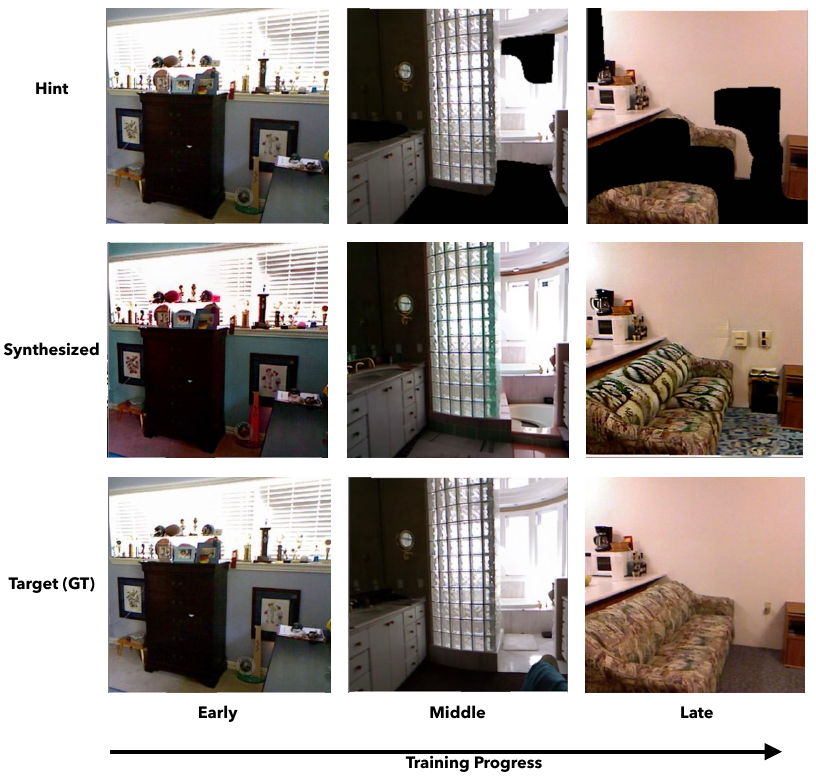}
  \vspace{-10pt}
  \caption{We condition our renderer with texture hints (first row), copying texture from the GT target image for a random subset of primitives. The blacked-out regions are areas the network must inpaint conditioned on the surrounding texture, text description, and primitives. Early in training, significant color shifts are present for the non-inpainted regions; these unwanted shifts get suppressed as training progresses.}
\label{fig:texture}
\end{figure}

\begin{figure}
  \includegraphics[width=\linewidth]{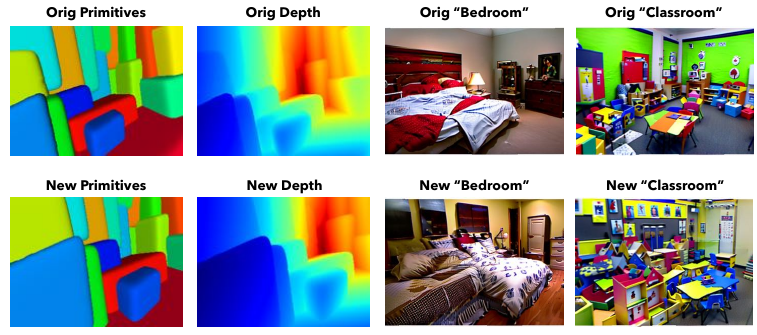}
  \vspace{-10pt}
  \caption{Additional camera move examples. Notice how the 3D world represented by the primitives is respected in the generated images.} 
\label{fig:cam2}
\end{figure}

\begin{figure*}
  \includegraphics[width=\textwidth]{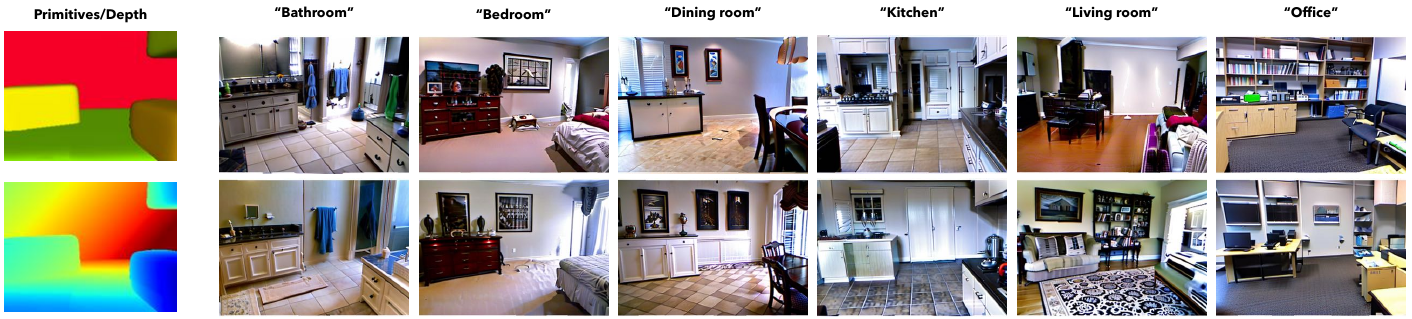}
  \vspace{-15pt}
  \caption{Additional examples demonstrating user-generated primitives and synthesized results with different scene labels. 
  }
\label{fig:author2}
\end{figure*}

\paragraph{Limitations:} It is natural to try and compel Blocks2World to produce objects of a known type ({\tt fridge}, say).
However, the alignment between our primitive representation and a semantic segmentation of the image is
too poor for this to work currently; typically, the method will produce an image with a {\tt fridge} in it,
but not in the right place (Fig~\ref{fig:limit_a}).  Diffusion models can cope with eccentric configurations of primitives for a given scene type, but tend to produce scenes that are curious when inspected in detail (cf the unusually sociable arrangement of commodes in Fig. 7, "medieval bathroom").  Fixing this behavior would require much deeper scene understanding than is currently available. The primitive representation is a compromise between
simplifying interaction and controlling detail.  It is easy to interact with a representation in
terms of simple primitives, but detailed control is compromised;  
if one really wishes to control the position of, say, a finial on a bedpost, our primitive decomposition is
not going to work.

\bibliographystyle{ieee_fullname}
\bibliography{sample-base, anand_ref}

\end{document}